\documentclass[default, iicol]{sn-jnl}

\jyear{2021}%

\theoremstyle{thmstyleone}%
\theoremstyle{thmstyletwo}%

\theoremstyle{thmstylethree}%

\usepackage{longtable}
\usepackage{lscape}
\usepackage{caption}
\usepackage{subcaption}

\begin{document}

\title[\hspace{13.7cm} Distributed Reinforcement Learning for Robot Teams: A Review]{Distributed Reinforcement Learning for Robot Teams: A Review}

\author[1]{\fnm{Yutong} \sur{Wang}}\email{e0576114@u.nus.edu}

\author[1]{\fnm{Mehul} \sur{Damani}}\email{damanimehul24@gmail.com}

\author[1]{\fnm{Pamela} \sur{Wang}}\email{wangyeelin@gmail.com}

\author[1]{\fnm{Yuhong} \sur{Cao}}\email{caoyuhong@u.nus.edu}

\author*[1]{\fnm{Guillaume} \sur{Sartoretti}}\email{guillaume.sartoretti@nus.edu.sg}

\affil*[1]{\orgdiv{Department of Mechanical Engineering}, \orgname{National University of Singapore}, \orgaddress{\street{\\9 Engineering Dr 1}, \postcode{117575}, \country{Singapore}}}

\abstract{ 

\textbf{Purpose of review}: Recent advances in sensing, actuation, and computation have opened the door to multi-robot systems consisting of hundreds/thousands of robots, with promising applications to automated manufacturing, disaster relief, harvesting, last-mile delivery, port/airport operations, or search and rescue. The community has leveraged model-free multi-agent reinforcement learning (MARL) to devise efficient, scalable controllers for multi-robot systems (MRS). This review aims to provide an analysis of the state-of-the-art in distributed MARL for multi-robot cooperation.

\textbf{Recent findings}: Decentralized MRS face fundamental challenges, such as non-stationarity and partial observability. Building upon the ``centralized training, decentralized execution'' paradigm, recent MARL approaches include independent learning, centralized critic, value decomposition, and communication learning approaches. Cooperative behaviors are demonstrated through AI benchmarks and fundamental real-world robotic capabilities such as multi-robot motion/path planning.

\textbf{Summary}: This survey reports the challenges surrounding decentralized model-free MARL for multi-robot cooperation and existing classes of approaches. We present benchmarks and robotic applications along with a discussion on current open avenues for research.

}

\keywords{Multi-Robot Systems; Reinforcement Learning; Cooperation; Communication Learning; Mixed cooperative-competitive Settings; Motion Planning \vspace{-0.3cm}}

\maketitle

$\,$ \vspace{-0.875cm}
\section{Introduction}
\label{sec:introduction}

With the recent advances in sensing, actuation, and computation, we are quickly approaching a point at which real-life applications will involve the deployment of hundreds, if not thousands, of robots, for tasks such as automated manufacturing and deliveries~\cite{nagele2019multi}, environmental monitoring and exploration~\cite{DBLP:conf/dars/MaMLS16}, disaster relief/search and rescue~\cite{DBLP:conf/icinfa/WangZSP17}.
However, as the number of agents (i.e., robots) in the system grows, so does the combinatorial complexity of coordinating them.
Existing conventional methods such as handcrafted controllers struggle at keeping up with this increased need for performance, scalability, and flexibility, since they usually rely on the intuition/experience of their designer and on domain-specific knowledge.
Building upon the success of (single-agent) reinforcement learning (RL), the community has started to look to machine learning methods -- and in particular multi-agent reinforcement learning (MARL) -- to devise controllers for multi-robot systems, with a particular focus on cooperation among robots and scalability to larger teams.

In this survey, we first introduce RL and MARL, as well as the different reward and training setups.
We discuss the challenges associated with MARL and MRS (Section~\ref{sec:background}), namely non-stationarity during training, partial observability, scalability, as well as challenges associated with communications (and, in particular, communication learning) among agents.
We then detail the main research directions that have been proposed recently to tackle these challenges and improve agent cooperation (Section~\ref{sec:cooperation}).
In particular, recent advances have often adopted the ``Centralized Training, Decentralized Execution'' (CTDE) paradigm, where agents rely on global information during training to boost cooperation, while still learning policies that only rely on local information and interactions among agents at execution time, thus improving scalability.
We report relevant approaches to decentralized multi-agent cooperation, namely independent learning, centralized critic, value decomposition, and communication learning methods.
We then discuss how the fundamental agent capabilities developed by these works can be assessed and visualized, both through AI benchmarks and real-world robotic tasks (Section~\ref{sec:applications}).
Most experimental demonstrations of MARL cooperation methods on robot consider motion/path planning problems, both as a fundamental multi-robot problem and because many robot teams consider mobile robots, whose (motion) coordination is of primary concern.
Finally, we take on a more critical view to describe open challenges in MARL, identifying and motivating future research in these areas (Section~\ref{sec:open-avenues}), and conclude this review in Section~\ref{sec:conclusion}.

We note that our review is by no means exhaustive.
For further information about the recent work in distributed MARL, we refer the reader to a set of excellent resources in this area~\cite{DBLP:journals/corr/abs-1908-03963, DBLP:journals/aamas/Hernandez-LealK19, DBLP:journals/tcyb/NguyenNN20, DBLP:journals/air/GronauerD22,
DBLP:journals/corr/abs-1906-04737}.
Furthermore, and more generally for multi-agent systems (MAS), Jorge and Magnus discussed a number of decentralized control and coordination strategies~\cite{cortes2017coordinated}.
Elio et al.~\cite{DBLP:journals/firai/TuciAA18} and Zhi et al.~\cite{DBLP:journals/arc/FengHSS20} reviewed recent advancements in multi-robot cooperative object transport and collaborative manipulation, respectively.

\section{Background}
\label{sec:background}

\subsection{Single-Agent Reinforcement Learning}

Reinforcement Learning (RL) addresses the fundamental problem of sequencing intelligent decisions in an environment with discrete time-steps to maximize long-term cumulative return, and is often formalized as a Markov Decision Process (MDP).
An MDP is defined by a tuple $(S,A,R,T,O,Z,\rho,\gamma)$, where $S$ is the state space, $A$ the action space, $R:S\times A\times S \to \mathbb{R}$ the reward function, defining the agent's reward at each time-step.
$T:S\times A\times S \to [0,1]$ is the state transition function, defining the probability of reaching state $s'$ by taking action $a$ in state $s$. $O: S\times A\to Z$ is the observation function, from which the agent's observation (sometimes referred to as its \textit{state}) $z \in Z$ is sampled at each time-step. $\rho(s_0)$ is the distribution of initial states, and $\gamma \in [0,1]$ is the \textit{discount factor}, setting the trade-off between immediate and future rewards.

The goal of an agent is to learn an optimal decision strategy, more commonly referred to as a \textit{policy}, $\pi^*: S \to A$, which produces a distribution over possible actions in each state, such that the long-term, cumulative discounted return is maximized:
\vspace{-0.2cm}
\begin{equation} \label{eq1}
    \pi^* = P(a \mid s) = \underset{\pi \in \Pi}{\arg\max} \,\,\, \mathbb{E}_{s\sim\rho(s_0)} \left[ V_\pi(s) \right],
\end{equation}
where $\Pi$ is the set of all policies and $V_\pi(s)$ is the \textit{state value} function, i.e., the expected return when starting in state $s$ and following policy $\pi$:
\vspace{-0.4cm}
\begin{equation}
    V_\pi(s) = \mathbb{E}_{\pi,T}\Big[ \sum_{t=0}^{\infty} \gamma^t \, R(s_t,a_t,s_{t+1}) \mid s_0=s \Big].
\end{equation}
Similar to $V_\pi(s)$ , $Q^{\pi}(s,a)$ is defined as the \textit{state-action value} function, i.e., the expected return when starting in state $s$, taking action $a$ and thereafter following policy $\pi$:
\vspace{-0.1cm}
\begin{equation}
        Q_\pi(s,a) = \mathbb{E}_{\pi,T}\Bigg[ \sum_{t=0}^{\infty} \gamma^t \, R(s_t,a_t,s_{t+1}) \, \Bigl\lvert \,\, \begin{matrix} s_0=s \\ a_0=a \end{matrix} \Bigg].
\end{equation}

\noindent There are two broad classes of methods to learn an optimal policy: value-based and policy-based methods.

Value-based methods aim to estimate the optimal policy's corresponding state value function $V^*(\textbf{s})$ or state-action value function $Q^*(\textbf{s, a})$.
Once such a function is obtained, the optimal policy is obtained by acting greedily with respect to it.
These functions can also be computed through tabular methods using algorithms such as value iteration and policy iteration, if the transition model is known.
However, a transition model is generally not available in most realistic environments, and algorithms have to rely on sampling to approximate these functions.
This can either be done by relying on full rollouts through the environment (Monte-Carlo methods), or via bootstrapping.
DQN is a very popular sampling-based algorithm that relies on bootstrapping to learn the optimal action-value function, which is parameterized using a deep neural network~\cite{DBLP:journals/corr/MnihKSGAWR13}.

In contrast to value-based methods, policy-based methods directly try to learn the optimal policy $\pi^*$.
Typical policy-based setups involve a parameterized policy $\pi_{\theta}$, whose parameters are optimized to maximize the objective in Eq.~\eqref{eq1} using either gradient-based or gradient-free approaches.
The recent advent of deep learning has made it possible to represent powerful non-linear policies through a deep neural network and optimize the network weights through gradient-based methods.
REINFORCE is one of the most fundamental and popular policy gradient algorithms which uses actual returns from a trajectory to compute such weights gradients~\cite{DBLP:conf/nips/SuttonMSM99}.
However, since it relies on noisy returns for optimization, REINFORCE generally suffers from high variance, drastically slowing down its convergence.

Actor-critic methods are a special class of policy-based methods which learn both an explicit policy representation and a state-value function for that policy.
Instead of direct (often noisy) returns from the environment like in REINFORCE, the learning of the policy (actor) is guided by the value function (critic).
In practice, this significantly reduces variance and stabilizes training, but at the cost of introducing some bias, which is generally an acceptable trade-off.
Actor critic methods have grown massively in popularity since their introduction and include most state-of-the-art RL algorithms, such as A3C~\cite{DBLP:conf/icml/MnihBMGLHSK16}, PPO~\cite{DBLP:journals/corr/SchulmanWDRK17}, SAC~\cite{DBLP:conf/icml/HaarnojaZAL18}, and DDPG~\cite{DBLP:journals/corr/LillicrapHPHETS15}

This section is meant to provide a basic background in RL and is in no way comprehensive.
For more detailed information, we refer the reader to a set of excellent resources in this area~\cite{sutton2018reinforcement,DBLP:journals/spm/ArulkumaranDBB17,DBLP:journals/jair/KaelblingLM96,DBLP:journals/ftml/Francois-LavetH18}.

\subsection{Multi Agent Reinforcement Learning}

MARL generalizes the single-agent MDP to a Markov game, which can be described by a tuple $(N,S,A,R,T,O,Z,\rho,\gamma)$ where $N$ is the number of agents, $S$ the state space, \textbf{$A$} the joint action space, defined as $\textbf{A}=A_1\times A_2\times....\times A_N$.
The reward function $R:S\times \textbf{A}\times S \to \mathbb{R}^n$ computes $N$ rewards $[r_1,r_2....,r_N]$ at each time-step, one for each agent.
$T:S\times \textbf{A}\times S \to [0,1]$ is the transition function, which defines the probability of reaching state $s'$ after taking joint action $\textbf{a}=[a_1,a_2,...,a_n]$ in state $s$.
$\textbf{O}=[O_1,O_2,...,O_N]$ are the set of observation functions, from which each agent's observation $O_i(s): S\times \textbf{A}\to Z_i$ is sampled at each time-step, where $\textbf{Z}=[Z_1,Z_2,...,Z_N]$ are the agents' observation spaces.
Finally, $\rho(s_0)$ and $\gamma$ are the initial state distribution and discount factor respectively.

The above formulation assumes that the multi-agent system is \textit{heterogeneous}, i.e., each agent can have a distinct action space and observation function.
On the other hand, \textit{homogeneous} systems assume that agents have the same action space and observation function, i.e., $A_1=A_2=...A_N$, $O_1=O_2=...=O_N$ and $Z_1=Z_2=...=Z_N$. 
However, agents in a homogeneous system can still have differing policies that allow them to perform specialized roles. 
When specialization is not a requirement, homogeneous systems can be extended by sharing the same policy between agents i.e., having homogeneous policies, which allows for efficient and more scalable learning by relying on parameter sharing~\cite{DBLP:conf/atal/GuptaEK17}.

MARL setups can be broadly classified by their reward functions, which can induce cooperative or competitive behaviors, as well as on the basis of their learning setups, which can drastically influence the type and quality of policies learned.
These classifications are briefly described in the following sections.

\subsection{Reward Setup}
\label{sec:rewardsetup}

\textbf{Fully Cooperative}: In fully cooperative settings, all agents share the same reward. That is, $r_1=r_2=...=r_N$.
These settings face the challenge of credit assignment, as individual agents are not rewarded only on the basis of the action they took, but instead based on the joint action of the entire team.
This leads to a high variance in rewards, which is particularly significant during early stages of training, where many exploratory actions are being taken.
Alleviating this variance requires the use of methods capable of either implicit or explicit \textit{credit assignment}.
Common environments in this setting include the suite of StarCraft Micromanagement environments~\cite{DBLP:conf/atal/SamvelyanRWFNRH19} and many Multi-Agent Particle environments~\cite{DBLP:conf/aaai/MordatchA18}. 

\textbf{Competitive}: 
In competitive settings, agents receive and try to selfishly maximize their own local rewards, which are often conflicting in nature to other agents. 
That is, agents have conflicting goals and a high reward for some agents usually comes at the cost of low rewards for the other ones.
Zero-sum games are a characteristic example of two player fully competitive environments, where the sum of all agent rewards is equal to $0$.
Examples of two-player zero-sum game environments include Chess/Shogi/Go~\cite{DBLP:journals/corr/abs-1712-01815} and Pong~\cite{DBLP:journals/nature/MnihKSRVBGRFOPB15,DBLP:journals/corr/TampuuMKKKAAV15}.
Larger competitive benchmarks for MARL include Pommerman~\cite{DBLP:conf/aiide/ResnickEHBFTCB18} and Neural-MMO~\cite{DBLP:journals/corr/abs-1903-00784}.
It is also possible to have a hybrid setting, which consists of cooperative multi-agent teams in competition with other teams. 
That is, the reward setup is fully cooperative at the team level but competitive at the game level.
An example of such an environment is the prey-predator game, where a team of pursuer agents is tasked with cooperatively capturing/caging evader agents \cite{DBLP:conf/iclr/KimPS21, DBLP:conf/iclr/KimMHKLSY19}.

\textbf{Mixed Cooperative-Competitive}: 
In mixed environments, agents receive their own local (often shaped) rewards.
These rewards, although given with the ultimate objective of maximizing a global metric or the team reward (sum of all local rewards), might induce selfish/undesirable behaviors in agents as they don't always directly align with the overall objective.
Mixed environments lie at a unique intersection where they face a trade-off between efficient learning and optimality, i.e., shaped rewards accelerate learning but induce selfish behaviors which affect optimality.
Examples of such settings are multi-robot path planning (Section~\ref{MAPF}) or multi-agent traffic signal control (MATSC)~\cite{DBLP:journals/tits/ChuWCL20}, where agents are locally rewarded for reaching a target location and minimizing the local traffic queue length respectively.
However, the performance of these systems as a whole is judged on global metrics - the throughput (agents reaching their goals per time-step) and average travel time of vehicles respectively. \\[-0.3cm]

\noindent \textbf{This work focuses on robot teams in both fully cooperative and mixed cooperative-competitive settings.}

\subsection{Learning Setup}
\label{learning-setup}

\subsubsection{Centralized Learning}

Centralized training models the multi-agent system as a single-agent and aims to learn a joint action from the joint observations of all agents.
Centralized methods operate on the assumption that information can be centralized, both during training and execution, something which is often not achievable in practice.
Additionally, centralized learners encounter a combinatorially growing action and observation space as the number of agents in the system grow.
This makes optimization hard and introduces issues such as the lazy agent problem~\cite{DBLP:conf/atal/SunehagLGCZJLSL18}.
In conclusion, although they are convenient to model and enjoy access to all available environment information through centralization, centralized learners are both severely limited in scalability and difficult to optimize.

\subsubsection{Independent Learning} 

In independent learning, each agent independently learns its own policy based on its own observation while considering other agents as a part of the environment.
As a result, independent learning can be decentralized in both training and execution, allowing it to scale considerably more than centralized methods.
However, by considering other agents as a part of the environment, independent learners forgo the Markov property and learn in an environment that is inherently non-stationary due to the changing policies of other agents.
Additionally, in contrast to centralized learning, independent learners have limited observability and are unable to enjoy any gains which come from information centralization.

\subsubsection{Centralized Training Decentralized Execution (CTDE)}
\label{sec:CTDE}

The CTDE paradigm aims to get the best of both centralized and independent learning paradigms, by allowing agents to learn in a centralized setting while executing policies in a decentralized setting.
This allows agents to benefit from additional information available during training from centralization, as well as enjoy the scalability benefits of decentralized execution.
Additional information can allow the learning of centralized critics~\cite{DBLP:conf/nips/LoweWTHAM17,DBLP:conf/aaai/FoersterFANW18} and value factorization networks~\cite{DBLP:conf/icml/RashidSWFFW18,DBLP:conf/atal/SunehagLGCZJLSL18}, as well as allow techniques such as parameter sharing (sharing network parameters between different agents) to accelerate learning~\cite{DBLP:conf/atal/GuptaEK17}.
These works are discussed in more detail in Section~\ref{sec:applications}.
The CTDE regime has gained remarkable popularity in MARL, and is particularly desirable in settings where a simulator is available for learning, as is often the case for robotic tasks. \\[-0.2cm]

\noindent \textbf{This work primarily focuses on the CTDE paradigm but also discusses independent learning in some detail.}

\subsection{Challenges of MARL}
\label{challeges}

\subsubsection{Non-Stationarity}

\begin{figure}[t]
\includegraphics[width=\linewidth]{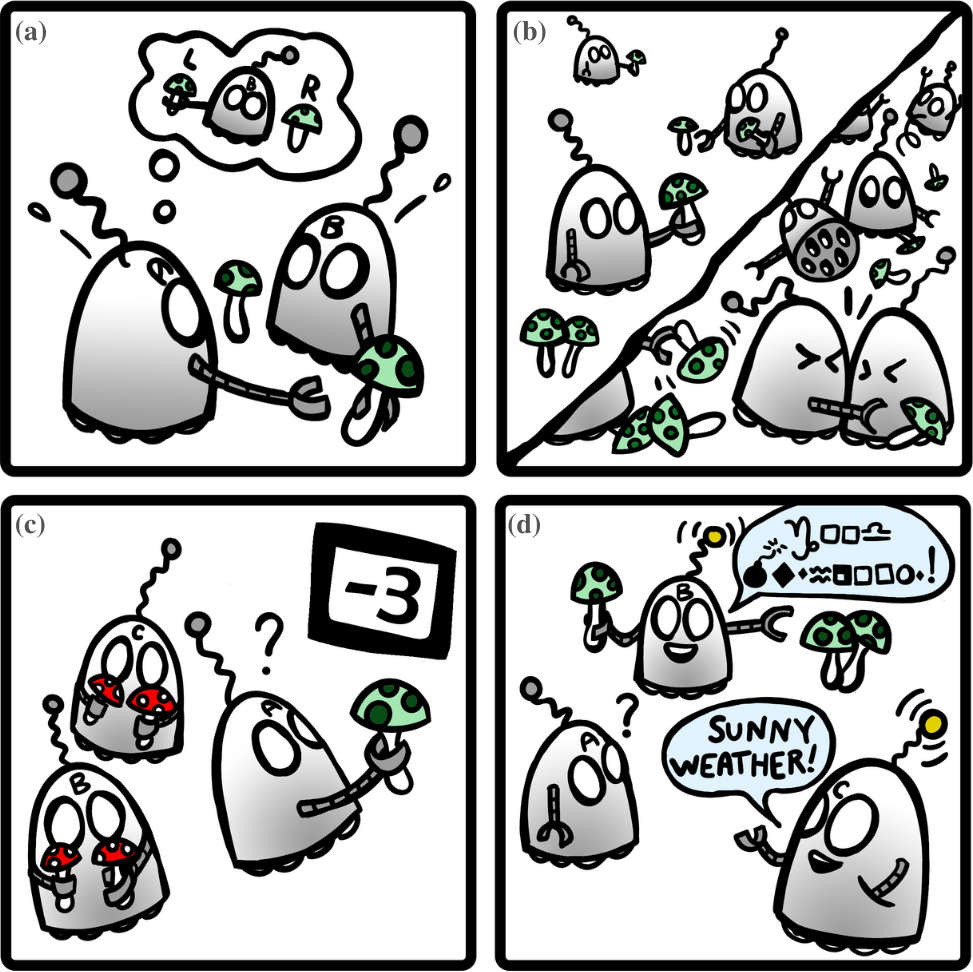}
\caption{\textbf{Key challenges in MARL: (a)} \textit{Non-Stationarity} - Agents learn in an ever-changing environment, where other agents also constantly update their behavior. As a result, predictions based on past experiences may not be accurate anymore. Here, agent A (wrongly) predicts that agent B will go for the left mushroom.
\textbf{(b)} \textit{Scalability} - Agents trained in a smaller team may struggle at generalizing their strategy to a larger one.
\textbf{(c)} \textit{Partial Observability} - Agents may be confused by shared rewards, which often depend on the action of agents beyond their sensing range.
\textbf{(d)} \textit{Communication} - Agents need to identify and encode relevant information in a commonly agreed upon manner, and must often learn to select whom to speak/listen to, to avoid being overwhelmed. Here, agent A cannot understand the useful message from agent B, deciding instead to listen to the useless, yet understandable message from agent C.
Image by and courtesy of Anna Sz{\"u}cs.
}
\label{fig:challenges}
\end{figure}

From the perspective of a training agent, which considers other agents to be a part of the environment, a multi-agent environment is inherently \textit{non-stationary} due to the changing policies of those other agents who are also learning at the same time. 
In other words, the transition function $T$ is non-stationary from the perspective of a single learning agent that cannot observe the policies of other agents.
This poses a complex learning problem, where the evolving policies of different agents interact with each other in an unpredictable manner and often lead to undesirable outcomes such as oscillating policies (Fig.~\ref{fig:challenges} (a)).
Convergence guarantees, which are well-defined in stationary environments (most single-agent environments), are not applicable in non-stationary environments as the Markov property does not hold there.
Additionally, learning in a non-stationary environment implies that data collected in the past was governed by a different environment dynamics (i.e., ``off-environment'') function based on the past policies of other agents.
This affects the performance of all algorithms that maintain an experience buffer to learn from past data.

\subsubsection{Scalability} 

Scalability is a multi-faceted challenge, which both encompasses training policies that can scale, i.e., generalize to smaller/larger teams, and algorithms that can handle the training of large teams, without a considerable effect on performance and (computational) cost (Fig.~\ref{fig:challenges} (b)).
Both kinds of scalability are crucial in the real world, where it is common for robots to be deployed on tasks of varying scales and expensive for robots to be retrained due to the changes in the team size (such as agent addition or breakdowns).
However, most algorithms face a trade-off between non-stationarity and scalability.
Non-stationarity can be alleviated by centralization, but centralization faces scalability challenges such as an exponentially growing state and action spaces (in particular, most algorithms that rely on communications see the number of messages transmitted scale exponentially) and the high cost of centralizing information.
Similarly, independent learning can scale up, but faces growing non-stationarity as more learning agents are added to the system.
The CTDE paradigm (Section~\ref{sec:CTDE}) balances these trade-offs reasonably well.
However, training functions such as a centralized critic or a value decomposition network is also limited as an extremely large network can be difficult to optimize.
Out of the three learning setups, CTDE-based algorithms have the most potential to further mature and increase scalability by incorporating techniques such as parameter sharing~\cite{DBLP:conf/atal/GuptaEK17}.

\subsubsection{Partial Observability}

In many environments, agents are required to learn a policy based on limited (often noisy) information about the environment, i.e, partially observable settings.
This is even more prominent in real-world applications where cost and characteristics of robot sensors can considerably restrict the information available to a robot.
For example, a forward facing camera cannot provide information about the other directions around a robot.
Partial observability is an even more profound challenge in multi-agent environments as an agent might not be able to observe other agents in the system.
This exacerbates non-stationarity and makes learning extremely hard, especially in fully cooperative settings where an agent might not be able to infer why it received a low reward, because the agent responsible for the low reward is not currently observable (Fig.~\ref{fig:challenges} (c)).
Partial observability can be addressed through network architectures that incorporate memory~\cite{DBLP:conf/aaaifs/HausknechtS15}.

\subsubsection{Communication}

Key issues such as non-stationarity and partial observability can be mitigated by allowing agents to communicate information within the team, to augment each others' knowledge/observations~\cite{DBLP:conf/nips/FoersterAFW16,DBLP:conf/nips/SukhbaatarSF16,DBLP:conf/iclr/SinghJS19}.
However, employing predefined/handcrafted communication protocols generally restricts the benefits and flexibility of communication, by relying greatly on the intuition of its designer and/or on domain knowledge.
On the other hand, allowing agents to learn what to share and how to encode this information, as well as whom to communicate/listen to, comes with fundamental challenges (Fig.~\ref{fig:challenges} (d)), which are being addressed by the \textit{communication learning} community (see Section~\ref{sec:comms-learning}).
First, allowing agents to identify and encode relevant information into an efficient message is the first challenge, and there is currently no conclusive evidence on what information is required for agents to learn an optimal policy.
Second, the choice/learning of a high-quality communication topology, which can accurately capture the intricate relationships between agents while meeting real-world communication constraints (e.g., bandwidth or range), is also nontrivial.
Finally, scalability remains one of the hardest challenges in communication, where larger team sizes can see agents overwhelmed with (potentially contradictory) messages, essentially introducing noise in the agents' observations and leading to poor performances (i.e., the \textit{chatter} problem).

\section{Principles of Cooperation} 
\label{sec:cooperation}

\begin{figure}
\includegraphics[width=\linewidth]{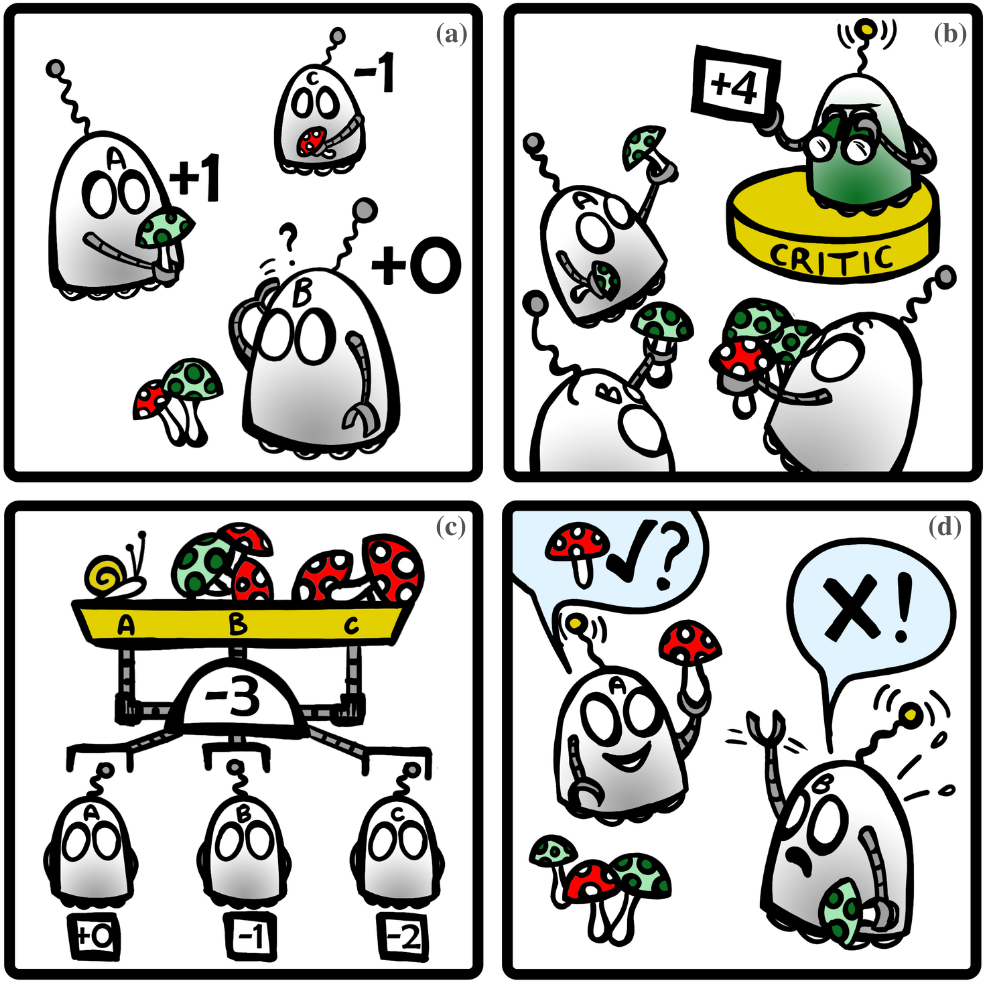}
\caption{\textbf{Cooperation approaches in MARL. (a)} \textit{Independent Learning} - Agents learn policies (here, from individual rewards) by treating other agents as part of their common environment.
\textbf{(b)} During training, a \textit{Centralized Critic} can provide a more accurate cooperative baseline - the state value, i.e., the expected long-term return from the current state, by relying on augmented state and policy information from all agents. Learned policies remain decentralized.
\textbf{(c)} \textit{Factorized Value Functions} - Agents learn to explicitly address the \textit{credit assignment} problem to transform a shared reward into individual contributions that can be used to update their individual policy,
\textbf{(d)} \textit{Communication Learning} - Agents learn to identify, encode, and share relevant information to augment each other's knowledge about the system.\\
Image by and courtesy of Anna Sz{\"u}cs.
}
\label{fig:AI_cooperation}
\end{figure}

This section describes the four classes of approaches to mitigate the challenges discussed in Section~\ref{sec:background} and encourage agent cooperation.
These approaches are illustrated in Fig.~\ref{fig:AI_cooperation}, while Table~\ref{table:allcoop} summarizes the main representative cooperation algorithms reviewed in this section.

\subsection{Independent Learners}

Independent learners learn their own policy based on their own observation using either shared (fully cooperative) or local (mixed cooperative-competitive) rewards while considering other agents as a part of the environment.
While vanilla independent learning often functions well in practice~\cite{DBLP:journals/corr/TampuuMKKKAAV15}, more recent independent learning works have focused on developing methods that address non-stationarity in MARL environments, which is particularly significant in fully cooperative settings where agents receive a joint team reward and an agent's action being individually-optimal is not enough to ensure team-level optimality.
Lauer et al. introduced an optimistic Q-Learning algorithm, which only updates Q-values when there is a guaranteed improvement and ignores negative updates by attributing low returns to sub-optimal actions of other agents~\cite{DBLP:conf/icml/LauerR00}.
Softer versions of this were introduced with the aim of ultimately converging to accurate action values~\cite{DBLP:conf/iros/MatignonLF07, DBLP:conf/atal/PanaitSL06}.
Matignon et al.~\cite{DBLP:conf/iros/MatignonLF07} used a constant smaller learning rate for negative updates, while Panait et al.~\cite{DBLP:conf/atal/PanaitSL06} used a decaying temperature to initially show leniency in negative updates and progressively decrease leniency as a state-action pair's visit count increases.
More generally, there is a spectrum of independent Q-Learning methods ranging from totally optimistic~\cite{DBLP:conf/icml/LauerR00}, moderately optimistic~\cite{DBLP:conf/iros/MatignonLF07,DBLP:conf/atal/PanaitSL06} and vanilla Q-Learning.
Although originally introduced for tabular settings, these methods have been successfully extended to deep RL settings~\cite{DBLP:conf/atal/PalmerTBS18,DBLP:conf/icml/OmidshafieiPAHV17}.
While the methods described above focus on fully cooperative settings, using intrinsic rewards that encourage social influence over other agents has also proven effective in mixed cooperative-competitive settings~\cite{DBLP:conf/icml/JaquesLHGOSLF19}. 
In addition to off-policy methods, which are more popular in MARL due to their sample-efficiency, recent methods utilizing state-of-the-art on-policy algorithms such as PPO have also shown great promise~\cite{DBLP:journals/corr/abs-2202-00082,DBLP:journals/corr/abs-2103-01955} 

While optimistic Q-Learning methods can address non-stationarity to a certain extent, many off-policy deep RL algorithms introduce another form of non-stationarity through their use of an experience buffer.
In order to stabilize training and efficiently reuse data, these algorithms maintain a buffer which stores past environment transitions.
However, in the case of a multi-agent environment, the experience buffer consists of off-environment, possibly obsolete data that can introduce training instabilities.
Given the central role played by experience replay, Foerster et al. introduced two methods to stabilize experience replay in MARL~\cite{DBLP:conf/icml/FoersterNFATKW17}.
The first method introduced an importance sampling correction to decay updates from transitions where the joint policy had changed significantly.
The second method, which was partially inspired from hyper Q-Learning~\cite{DBLP:conf/nips/Tesauro03}, introduced a fingerprint as an input feature to disambiguate the age of the data and serve as an indicator for the quality of other agents' policies.
With a similar objective, Omidshafiei et al. introduced Concurrent Experience Replay Trajectories to sample concurrent experience from the replay buffers of different agents, thus introducing correlations in local policy updates~\cite{DBLP:conf/icml/OmidshafieiPAHV17}.

\subsection{Centralized Critic}

The CTDE paradigm has made it possible to train agents in a centralized manner, allowing for the development of methods that reduce non-stationarity and achieve better credit assignment.
Actor-critic algorithms are particularly well suited for the CTDE paradigm as the critic is required only during training and can thus be augmented with any desirable centralized information, such as the policies of all agents or any extra available state information.
This centralization of information during training can help alleviate the non-stationarity faced by independent learners.
In fully cooperative settings with a single joint reward, it is often sufficient to have a single centralized critic for all the agents, while in competitive and mixed cooperative-competitive environments with local agent rewards, each agent might be required to train its own critic.

Lowe et al. proposed a generally applicable algorithm called MADDPG for both cooperative and competitive settings, which learns a unique centralized critic for each agent that is conditioned on extra state information and other agents' actions~\cite{DBLP:conf/nips/LoweWTHAM17}.
Many works have proposed extensions to improve the performance and scalability of MADDPG and of centralized critics in general.
Iqbal et al. used attention to dynamically select relevant information for estimating the centralized critic's value function~\cite{DBLP:conf/icml/IqbalS19}.
ATT-MADDPG uses attention on the centralized critic to explicitly model the dynamic joint policy of teammates in order to improve cooperation~\cite{DBLP:conf/atal/MaoZXG19}.

In addition to non-stationarity, centralized critics can also be used for credit assignment in fully cooperative settings.
Foerster et al. proposed COMA, which uses a learned centralized critic to compute a counterfactual baseline for each agent that estimates the contribution of each agent to the shared reward~\cite{DBLP:conf/aaai/FoersterFANW18}.
Similarly, Zhou et al. proposed LICA, a framework for implicit credit assignment which directly ascends approximate joint action value gradients of the centralized critic, which is represented through a hyper-network to enable incorporation of latent state information directly into the policy gradients~\cite{DBLP:conf/nips/ZhouLSLC20}.

\subsection{Factorized Value Functions}

While centralized critics can guide policy learning efficiently in actor-critic methods, another way to utilize the CTDE paradigm effectively is to learn a team-value function and factorize it into agent-wise value functions that can be used during execution.
That is, each agent has its own parameterized Q-function, the learning of which is guided by the joint team value function.
The end-to-end framework allows for credit to be distributed implicitly from the centralized head to the individual learners, thus allowing for effective learning in joint reward settings.
In summary, centralized learning stabilizes the training, while value factorization ultimately ensures decentralized and scalable execution.

Value Decomposition Networks (VDN) factorize the joint team value function on the assumption that the joint function can be represented as a sum of individual value functions~\cite{DBLP:conf/atal/SunehagLGCZJLSL18}.
However, by enforcing the structural constraint of \textit{additivity}, VDN limits the class of functions that can be learned, i.e., not all possible factorizations are additive in nature.
QMIX enforces the constraint of \textit{monotonicity} between the team and individual value functions, which allows it to represent the team value function as a non-linear combination of individual agent values using a mixing network~\cite{DBLP:conf/icml/RashidSWFFW18}.
This enables it to represent a richer class of functions than VDN.
QTRAN aims to learn a more general factorization without any structural constraints by transforming the optimal value function into one which is easily factorizable and has the same optimal actions~\cite{DBLP:conf/icml/SonKKHY19}.
MAVEN extends QMIX and other value factorization methods by using a shared latent variable controlled by a hierarchical policy to guide committed and temporally extended exploration~\cite{DBLP:conf/nips/MahajanRSW19}.

\subsection{Communication Learning}
\label{sec:comms-learning}

Communication provides agents with the ability to share information and overcome many limitations in decentralized MARL.
For example, sharing observations/knowledge can allow agents to augment their knowledge, effectively alleviating partial observability.
Non-stationarity can also be mitigated by allowing agents to share their policies/actions (or intent) with each others, further boosting cooperation~\cite{DBLP:conf/iclr/KimPS21,DBLP:conf/nips/MahajanRSW19}.
Table~\ref{table:comm} summarizes the communication learning algorithms reviewed in this paper.

\subsubsection{Message Generation}

The first challenge in communication learning is learned message generation.
In the majority of existing works, agents learn to generate messages by encoding their current observation or the hidden state of a recurrent network fed with this observation.
Within a time step, messages are generated in one of two ways - by feeding each agents' information into a message generation module~\cite{DBLP:journals/corr/MaoGNLWKMSX17,DBLP:conf/iclr/KimPS21,DBLP:journals/corr/abs-2004-00470,DBLP:conf/nips/ZhangZL19,DBLP:conf/nips/JiangL18,DBLP:conf/iclr/JiangDHL20,DBLP:conf/icra/MaLM21,DBLP:conf/nips/SukhbaatarSF16,DBLP:conf/iclr/KimMHKLSY19,DBLP:conf/iclr/SinghJS19,DBLP:conf/aaai/LiuWHHC020,DBLP:journals/corr/abs-1712-07305,DBLP:conf/icml/DasGRBPRP19,DBLP:conf/corl/BlumenkampP20,DBLP:conf/atal/DuLMLRWCZ21,DBLP:journals/corr/abs-2202-03634,DBLP:conf/nips/FoersterAFW16,DBLP:journals/corr/abs-2002-05233}, or by allowing each agent to sequentially process a "message'' (or set of messages) passed among agents (following a given topology/topologies)~\cite{DBLP:journals/corr/PengYWYTLW17,DBLP:journals/ml/PesceM20,DBLP:journals/corr/abs-2201-11994}.

The generation of these message can be trained in two ways -  either via reinforced or differentiable communication learning~\cite{DBLP:conf/nips/FoersterAFW16}.
Reinforced communication learning implicitly trains message generation using the agent's own reward signal, naturally extending vanilla reinforcement learning.
Differentiable communication learning (DCL) explicitly trains message generation by backpropagating the learning gradients from the recipient agent(s) through the communication channel.
Because of this richer feedback, DCL is preferentially used, as it is generally able to handle more complex tasks.
However, DCL relies on the assumption of a differentiable communication channel, and thus cannot directly handle channels with unknown noise or binary/discrete messages.

Communications can be further enhanced by incorporating environmental or task information into their representation, and can even rely on the environment as the (implicit) communication channel.
Agarwal et al. proposed to incorporate the environment's intrinsic high-level structure by exchanging messages along the edges of a shared agent-entity graph, where the environment is described as a collection of discrete entities, and the positions of those entities serve as transmitted information~\cite{DBLP:conf/atal/AgarwalKSL20}.
Cao et al. found that grounding the communication channel such that messages are binding and verifiable allows agents to learn to communicate in mixed cooperative-competitive settings~\cite{DBLP:conf/iclr/CaoLLLTC18}.
Finally, ForMIC proposed to consider \textit{implicit} (stigmergic) communications among agents for multi-robot foraging, where robots create trails of pheromones in the environment that decay in concentration over time, to indicate the location and utility of distant food resources~\cite{DBLP:journals/ral/ShawWWS22}.

\subsubsection{Communication Topology}

Communication topologies can be classified into three types: predefined fixed topology~\cite{DBLP:journals/corr/MaoGNLWKMSX17,DBLP:conf/iclr/KimPS21,DBLP:conf/nips/SukhbaatarSF16,DBLP:journals/corr/PengYWYTLW17,DBLP:journals/corr/abs-1712-07305,DBLP:journals/ml/PesceM20,DBLP:journals/corr/abs-2201-11994}, predefined dynamic topology~\cite{DBLP:journals/corr/abs-2004-00470,DBLP:conf/nips/ZhangZL19,DBLP:conf/iclr/JiangDHL20,DBLP:conf/icra/MaLM21,DBLP:journals/corr/abs-2109-05413,DBLP:conf/atal/AgarwalKSL20,DBLP:conf/corl/BlumenkampP20} and learned topology~\cite{DBLP:conf/nips/JiangL18,DBLP:conf/iclr/KimMHKLSY19,DBLP:conf/iclr/SinghJS19,DBLP:conf/aaai/LiuWHHC020,DBLP:conf/icml/DasGRBPRP19,DBLP:journals/corr/abs-2202-03634}.
Predefined fixed topologies are manually defined before training based on prior assumptions and remain fixed throughout training and execution, e.g., global communications (complete graph among agents).
Similarly, predefined dynamic topologies rely on a pre-defined, non-learnable condition to decide whether communication would occur between any given pair of agents at every time step.
For example, many works have either assumed a fixed communication range around each agent~\cite{DBLP:journals/corr/abs-2004-00470,DBLP:conf/iclr/JiangDHL20,DBLP:conf/icra/MaLM21,DBLP:conf/atal/AgarwalKSL20,DBLP:conf/corl/BlumenkampP20}, a threshold on the confidence level of each agent’s individual decision~\cite{DBLP:conf/nips/ZhangZL19}, or the influence of the presence of other agents for decision adjustment~\cite{DBLP:journals/corr/abs-2109-05413}.

In contrast, learned topologies update the communication rule among agents via training, using gradient backpropagation similar to standard MARL.
The learned topology is usually dynamic, and the communication graph can change at every time step to match the needs of the task.
Common techniques for determining the subset of agents that should communicate include: hard attention~\cite{DBLP:conf/aaai/LiuWHHC020}, soft-max layers (i.e., sampling from a learned probability distribution)~\cite{DBLP:conf/iclr/SinghJS19}, and weight generator networks that select a fixed-sized subset of agents to communicate to~\cite{DBLP:conf/iclr/KimMHKLSY19}.
In practice, real-world constraints such as communication bandwidth need to be considered when applying communication topologies to robots, and some of these works specifically aim to achieve \textit{sparse} topologies~\cite{DBLP:conf/iclr/KimMHKLSY19}.
More recent work have further proposed to teach agents to minimize the amount of information shared at every time-step (i.e., select to generate shorter messages), allowing agents to even select when to avoid communication entirely~\cite{DBLP:conf/iros/FreedJSC20}.

\section{Towards Multi-Robot Applications}
\label{sec:applications}

\begin{figure*}
\includegraphics[width=\linewidth]{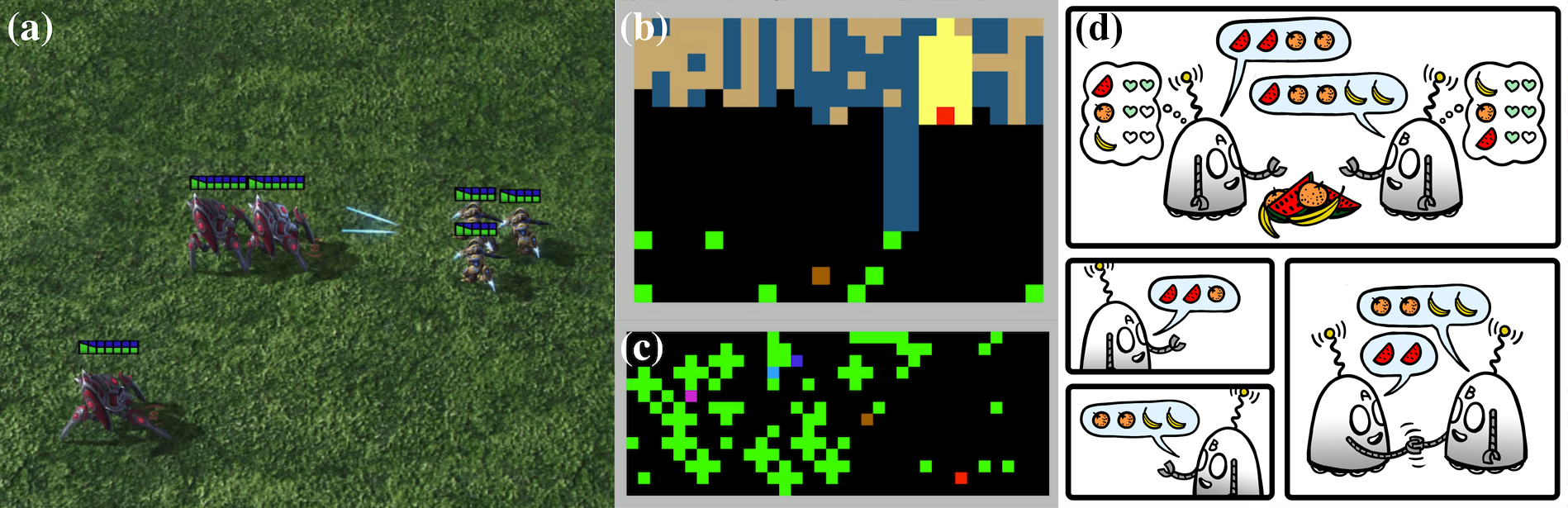}
\caption{\textbf{(a)} \textit{Starcraft Multi-Agent Challenge} example task (3s\_v\_3z), where the goal of the learning agents (left agents) is to eliminate all enemy agents (right agents), by selecting the correct movement/attack actions cooperatively.
\textbf{(b)} \textit{Cleanup} is a public goods dilemma where agents need to maximize the team's reward from individual, mixed cooperative-competitive rewards. Agents (red and brown squares) get rewards for consuming apples (green), which only (re-)grow when the river (blue) is clean from waste (brown), but do not directly get rewards for cleaning the river (yellow beam).
\textbf{(c)} Similarly, \textit{Harvest} is a common pool resource dilemma (tragedy of the commons), where apples (green squares) regrow at a rate proportional to the number of nearby apples, thus teaching agents (colored squares) to limit their (selfish) desires to consume apples, to reach a team-optimal behavior.
\textbf{(d)} \textit{Negotiation task}, where agents with individual needs have to divide up a set of items to maximize utility, based on successive communications.
Figure \textbf{(d)} by and courtesy of Anna Sz{\"u}cs.
}
\label{fig:AI_benchmarks}
\end{figure*}

In this section, we discuss AI benchmarks and real-world robotic tasks, which are used to assess the performance of cooperative multi-agent/-robot teams.
Simplified benchmarks allow us to test specific aspects of cooperation in controlled environments, and provide a common basis on which to compare different approaches.
In contrast, verifying MARL models on real robots in realistic scenarios helps to test the entire set of capabilities required for deployments, such as scalability and robustness to sensing/actuation noise.

\subsection{AI Cooperation Benchmarks}

Benchmarks serve as a common way to evaluate and compare the performance of different models.
We discuss cooperative, mixed cooperative-competitive, as well as communication learning benchmarks.

\subsubsection{Cooperative Setting Benchmarks}  
\label{sec:cooperative-benchmarks}

The most popular benchmarks for cooperative settings are the StarCraft II Multi-Agent Challenge (SMAC)~\cite{DBLP:conf/atal/SamvelyanRWFNRH19} and the multi-agent particle environment~\cite{DBLP:conf/aaai/MordatchA18} which both test the general ability of agents to work together in fully cooperative tasks with continuous observation and discrete action spaces.
SMAC focuses on team based cooperation by simulating a skirmish between two platoons, one of which is controlled by MARL agents and the other by the built-in, non-learning game AI.
An example of a symmetric, homogeneous battle scenario is shown in Fig \ref{fig:AI_benchmarks} (a), where the goal of the agents is to eliminate all adversaries to win the skirmish.
In order to make the tasks more challenging, SMAC includes asymmetric battle scenarios with heterogeneous team configurations, which require advanced tactics as well as the ability to find and act one's role in the team to ensure success.
The multi-agent particle environment is a simpler benchmark consisting of a multi-agent particle world with basic simulated physics~\cite{DBLP:conf/aaai/MordatchA18}.
This environment can be easily modified for the design of new tasks, in addition to the basic set of benchmark tasks that are provided such as prey-predator.

\subsubsection{Mixed Cooperative-Competitive Setting Benchmarks}

Mixed cooperative-competitive benchmarks focus on scenarios that threaten the stability of purely cooperative tasks, allowing us to test if agents will be able to cooperate, even in situations where their individual rewards motivates them to act in a way that might be detrimental to the collective good.
In particular, group dilemmas present agents with scenarios where they must work together to obtain higher team rewards, despite a built-in incentive to be selfish.
That is, if an agent does not cooperate with the team, they can maximize their own short-term return, but all agents in the environment (including themselves) will suffer the cost of their selfish actions in the long run.
The main benchmarks look at sequential social dilemmas, such as the common pool resource dilemma, and public goods dilemma which have been implemented in \textit{Cleanup}~\cite{DBLP:conf/atal/LeiboZLMG17} and \textit{Harvest} respectively.
In \textit{Cleanup} shown in Fig \ref{fig:AI_benchmarks} (b), agents gain reward from consuming apples, which only regrow when the river is clean from waste.
Although the team as a whole benefits from agents cleaning the river, these cleaning agents are not directly rewarded as they do it, while other agents are rewarded for consuming newly-regrown apples.
In \textit{Harvest} shown in Fig \ref{fig:AI_benchmarks} (c), apples regrow based on how many apples are nearby.
Harvest is a common-pool resource appropriation problem (\textit{tragedy of the commons)}, where agents directly gain reward from consuming apples, which regrow naturally proportionally to the amount of nearby apples.
Therefore, agents need to limit their selfish desire to consume as many apples as they can, to help maximize the regrowth rate and maximize the overall long-term team return.

Team return -- the overall sum of rewards received by all agents in the long run -- is traditionally used to track training progress in fully cooperative/competitive tasks.
However, team return is an insufficient metric in mixed cooperative-competitive settings, where improving agent cooperation does not necessarily immediately increase returns, and returns even often see a downwards trend before going back up once a nearer-team-optimal cooperation equilibrium is reached.
That is, unlike single agent cases where it is easier for the agent to self-correct to a more sustainable strategy, agents in a MAS often fall into socially-deficient Nash equilibria, where changing their own behavior is insufficient to change the overall team behavior, and thus acting greedily remains the dominant strategy~\cite{DBLP:conf/aaai/ClausB98}.
As a result, mixed cooperative-competitive benchmarks often turn to social outcome metrics~\cite{DBLP:conf/nips/PerolatLZBTG17}, such as equality and sustainability, which help shed light on the true progress of the cooperation learning process.

For a more quantitative evaluation, simple social dilemma matrix games, extended from game theory, have been used to analyze how different learning aids affect the probability of convergence to better/worse equilibria.
For example, general-sum coordination matrix games such as \textit{Stag Hunt} can help demonstrate how gifting between agents can improve the model's tendency to converge to the true, pro-social equilibrium~\cite{DBLP:conf/ijcai/WangBBLPS21}.

\subsubsection{Communication Benchmarks}

We examine communication learning benchmarks in both cooperative and mixed cooperative-competitive settings.
These benchmarks generally evaluate the agents' ability to share relevant information effectively, or to reach an agreement on how agents should compromise for the welfare of the team.
In purely cooperative settings, the most popular communication benchmarks are \textit{referential games}, where one agent is made aware of a target object, and has to generate a message that communicates to another agent which one to pick out of a set of candidate objects.
Referential games can be used to examine the viability of different mediums of communication~\cite{DBLP:journals/corr/abs-2106-02067}.
However, unlike non-communicative cooperative benchmarks, which mainly focus on the performance achieved within the game, referential games are also often used to measure the quality of the learned communication protocol.
For example, they can shed light into the ease of teaching such protocols to new agents, or, through the linguistic analysis of the compositionality of the learned language, can help examine the generalizability of communication protocols~\cite{DBLP:conf/nips/LiB19}.

Communication benchmarks in mixed cooperative-competitive setting aim to assess whether agents can learn to communicate successfully to reach a compromise for the good of the team, even when it does not necessarily benefit them individually.
The main benchmarks used are \textit{negotiation tasks}, where two agents with different hidden utility functions have to split a randomly initiated item pool by making successive proposals until both agents agree~\cite{DBLP:journals/corr/LewisYDPB17}.
These tasks can be used to examine the conditions required for agents to learn to reach a \textit{consensus} through sequential communications~\cite{DBLP:conf/iclr/CaoLLLTC18, DBLP:conf/atal/NoukhovitchLLC21}.

\subsection{Motion and Path Planning}
\label{MAPF}

Many robot teams are composed of mobile robots, of which controlling the (collision-free) motion is of primary concern.
More involved multi-robot tasks, such as RoboSoccer~\cite{DBLP:journals/corr/abs-2105-12196}, collaborative manipulation~\cite{DBLP:conf/atal/DingKMVNHR020}, multiple travelling salesman problem~\cite{DBLP:journals/corr/abs-2109-04205} and multi-UAV surveillance~\cite{DBLP:journals/tcom/HuZSSP20}, also consider the coordination/distribution of robots, and may add additional actions (e.g., kick the ball, or take measurements).
As a result,  multi-robot path/motion planning has become one of the most important and widely used tasks to evaluate the performance of deep RL methods for MRS.
In doing so, the hope is that the learned capabilities needed for cooperation in multi-robot path/motion planning will further extended to other, more general multi-robot tasks~\cite{DBLP:journals/corr/abs-2105-12196,DBLP:conf/atal/DingKMVNHR020,DBLP:journals/tcom/HuZSSP20}.
Additionally, thanks to the recent developments in robot motion control, it has become easier to verify deep RL algorithms for multi-robot motion/path planning on real robots~\cite{DBLP:journals/ijrr/FanLLP20,DBLP:conf/icra/XiaoHXA20}.

Many approaches have proposed to rely on independent learning for multi-robot path planning problems.
In particular, recently proposed approaches have built on both value-based and policy gradient-based RL algorithms for multi-robot path planning.
Wang et al. modified the dueling DQN algorithm and trained policies from images generated from robot-centric relative perspectives~\cite{DBLP:journals/ijon/WangDP20}.
Fan et al. proposed an approach that uses PPO, and allows agents to make decisions directly based on sensor-level inputs, which enables easier implementation on real robots~\cite{DBLP:journals/ijrr/FanLLP20}.
The learned policy is also integrated into a hybrid control framework to further improve its robustness and effectiveness.
Imitation learning has also been combined with deep RL to achieve better performance and sample efficiency for multi-robot path planning~\cite{DBLP:journals/ral/SartorettiKSWKK19, DBLP:journals/ral/DamaniLWS21}, by directly training agents to imitate the type of cooperative behaviors exhibited by centralized path planners, while showing scalability to thousands of agents.
Many of these works include experimental validation of the trained policies on hardware to verify their performance under realistic conditions~\cite{DBLP:journals/ral/SartorettiKSWKK19, DBLP:conf/icra/XiaoHXA20, DBLP:journals/ijrr/FanLLP20}.

More recent works in multi-robot path planning have relied on centralized critics and value factorization methods.
Notably, Marchesini et al. proposed a centralized state-value network to inject global state information in the value-based updates of the agents, demonstrating better performance than independent learning and VDN~\cite{DBLP:conf/iros/MarchesiniF21}.
Huang et al. relied on MADDPG with prioritized experience replay to efficiently utilize the collected experience~\cite{huang2020multi}.
He et al. applied MASAC based on cooperative waypoints search, and showed improvements in performance compared to other MARL algorithms such as MATD3 and MAAC~\cite{DBLP:journals/corr/abs-2112-06594}.
Value factorization approaches include de Witt et al., who introduced COMIX, which employs a decentralizable joint action-value function with a per-agent factorization, and showed significant improvement over MADDPG and VDN~\cite{DBLP:journals/corr/abs-2003-06709}.
Freed et al. proposed to rely on learned attention mechanisms to identify and decouple subsets of robots, which reduced the gradient estimator variances in a variety of motion/path planning tasks, showing improvements over independent learning and COMA~\cite{DBLP:journals/corr/abs-2112-12740}.

\begin{figure}
     \centering
     \includegraphics[width=\linewidth]{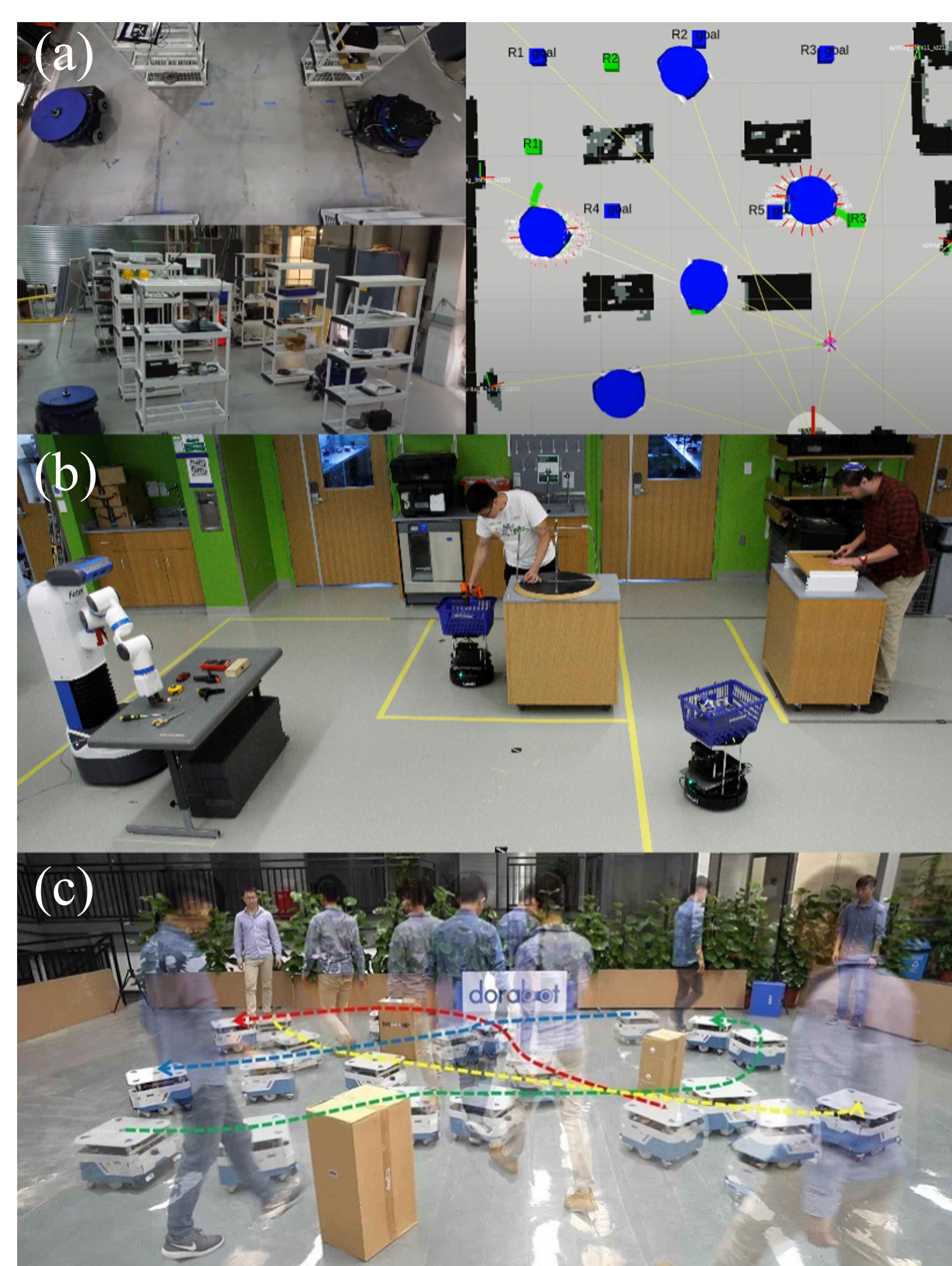}
    \caption{Experimental validation of deep MARL path/motion planning policies on MRS, considered in~\cite{DBLP:journals/ral/SartorettiKSWKK19}, \cite{DBLP:conf/icra/XiaoHXA20} and~\cite{DBLP:journals/ijrr/FanLLP20} respectively.
    Robots are assigned each a goal location and need to cooperatively plan collision-free paths to these goals.
    In \textbf{(a)} and \textbf{(b)}, robots need to avoid static obstacles in warehouse-like scenarios, while \textbf{(c)} considers crowded scenarios involving static and dynamic obstacles.
    \textbf{(b)} and \textbf{(c)} have been reused with the permission of the authors and copyright owners.
    }
    \label{fig:multi-robot path finding}
\end{figure}

\section{Open Avenues for Research}
\label{sec:open-avenues}

\textbf{Safe MARL}: When developing MARL algorithms intended to be directly implemented on robots in the real-world, safety is often of paramount importance.
Safe RL is defined as training an agent to learn an efficient policy, while minimizing violations of safety constraints, such as avoiding collisions with other robots or obstacles/humans, or ensuring the robot's safety/stability (e.g., enforcing actions that keep a flying robot in the air).
These safety guarantees need to be guaranteed when deploying the final learned policies, and sometimes even during training.
Numerous recent works have focused on safe single-agent RL.
They can be broadly classified into two types: modifying the optimization criterion and regulating the exploration process~\cite{DBLP:journals/jmlr/GarciaF15}.

However, the community is still in the early phases of transitioning from safe single-agent RL to safe MARL.
Ensuring safety becomes much more difficult in the presence of multiple agents, as the environment's instability increases, and solutions must now account for coupling among agents, particularly between those with competing interests.
In the limited research on safe MARL so far, we note Shalev et al., who devised a safe MARL algorithm for autonomous driving~\cite{DBLP:journals/corr/Shalev-ShwartzS16a}, and Zhang et al., who proposed a multi-agent algorithm that provably guarantees safety for arbitrary learned policies~\cite{DBLP:journals/corr/abs-1910-12639}.

\textbf{Simulation to Real-World}: Training in the real world is often costly, time-consuming, and sometimes dangerous or infeasible.
Thus, using simulated environments has become a common choice for robotics tasks, with a myriad of environments developed to simulate real-world tasks, such as MINOS (indoor 3D Navigation)~\cite{DBLP:journals/corr/abs-1712-03931}, Assistive Gym (physical human-robot interaction and robotic assistance)~\cite{DBLP:conf/icra/EricksonGKLK20}, SURREAL (robot manipulation)~\cite{DBLP:conf/corl/FanZZLZGCSF18}.
However, MARL research that successfully deploys policies trained in the virtual environment on physical robots has been rare so far~\cite{DBLP:journals/ijrr/FanLLP20,DBLP:journals/ral/FreedSC20}.
Due to complicated mechanical interactions, uncertainties in real system dynamics, time discretization, and numerous other challenges, building a simulator that perfectly reflects the real world is often impossible.
On the other hand, most MARL algorithms are notoriously sensitive to their input, and simple changes can drastically affect performance on the same task~\cite{DBLP:conf/aaai/0002IBPPM18}.
In particular, policies learned in low-precision simulations frequently cannot be transferred directly to real robots.
As a result, techniques that allow policies to be trained in simulation, and then adapted for implementation on robots in the real world, is an open question of importance in the community.

\textbf{Model-Based MARL}: Most of the algorithms introduced above are model-free methods, i.e., agents learn policy purely by interacting with the environment.
In contrast, model-based RL optimizes policies based on a known/learned model of the state-transition and/or rewards dynamics.
With the development of deep learning, model-free RL has attracted the most interest in recent year.
However, model-free RL requires large amounts of experiences to achieve acceptable performance, which, when applied on real robots, may render training times unreasonable and increase the risk of accidents and wear and tear on the hardware~\cite{DBLP:journals/jirs/PolydorosN17}.
In contrast, single-agent model-based RL methods have been proposed recently, which exhibit improved sample-efficiency and safety~\cite{DBLP:journals/trob/ThuruthelFRL19,DBLP:journals/ral/ThananjeyanBRLM20}.
Given these promising results, we believe that model-based MARL may be one of the next frontiers for MARL.
However, model-based MARL comes with a number of new challenges, as agents now need to learn the dynamics of an environment containing other agents, thus further increasing the impact of non-stationarity and partial-observability, as well as scalability.

\textbf{Decentralized Joint Decision-Making}: While existing communication learning methods can endow agents with the ability to augment each other's knowledge about the environment, share intent, or reach a consensus through negotiation, we believe that the holy grail of communication learning remains methods that can allow agents to \textit{consensually} generate joint plans from communications.
That is, advanced communications between agents may allow the team to reach centralized performances, while remaining fully decentralized.
In doing so, agents will most likely need to be able to diffuse relevant information in the team (including current knowledge and future intent/predictions), perform multiple back-and-forth communications to allow true dialogue to happen, and reach a consensus over joint actions to be selected for team-level optimality.
Although some of these capabilities have already been addressed by existing methods (as discussed in Section~\ref{sec:comms-learning}), their combination into a functioning, decentralized joint decision-making framework is not trivial.
In particular, we note that, while decentralized consensus-based methods have been used to achieve decentralized training in MARL~\cite{DBLP:journals/jzusc/ZhangYB21}, only a few works so far have truly addressed consensual joint action selection~\cite{DBLP:conf/icml/ZhangYL0B18}.

\section{Conclusions}
\label{sec:conclusion}

This survey provides a broad overview of recent work in distributed model-free MARL for MAS/MRS, and reports recent applications of MARL to AI and fundamental robotics tasks.
Cooperative MARL can be achieved by four general classes of approaches, namely independent learning, centralized critic, factorized value functions, and communication learning.
These methods have been investigated and validated extensively on AI benchmarks, but we also highlight recent works that were demonstrated on MRS, especially in the area of multi-robot motion/path planning.
Despite the exciting recent advances in the field, we also identify a few open avenues for future research in the application/improvement of MARL for multi-robot teams, such as safe MARL, sim2real transfer, model-based MARL, and decentralized joint decision-making.
We remain highly optimistic about the progress of the MARL community, and hope that this survey will inspire further developments in this very active and captivating field, to bring us ever closer to the wide-spread, safe implementation of these methods in the real-world to support the robotics deployments of tomorrow.

\section{Highlighted References to be added into the Reference List}
\label{sec:highlighted}

\noindent $\bullet$ Reference \cite{DBLP:conf/iclr/KimPS21}. This work allows globally-communicating agents to share intent by modeling the environment dynamics and other agents’ actions.

\noindent $\bullet\bullet$ Reference \cite{DBLP:conf/iclr/KimMHKLSY19}. This work allows agents to learn to estimate the importance of their observation/knowledge, to selectively broadcasts continuous messages to the whole team.

\noindent $\bullet\bullet$  Reference \cite{DBLP:conf/icml/JaquesLHGOSLF19}. This work proposed to encourage cooperation among agents by relying on an intrinsic reward that aims at maximizing their influence over each other.

\noindent $\bullet$ Reference \cite{DBLP:journals/corr/abs-2103-01955}.This work shows that independent learning using on-policy algorithms such as PPO can perform effectively in fully cooperative MARL environments.

\noindent $\bullet\bullet$  Reference \cite{DBLP:conf/icml/IqbalS19}. This work uses an attention mechanism in the centralized critic to dynamically select relevant information.

\noindent $\bullet$ Reference \cite{DBLP:conf/nips/ZhouLSLC20}.This work proposes a framework for implicit credit assignment which directly ascends approximate joint action value gradients of the centralized critic. 

\noindent $\bullet$ Reference \cite{DBLP:conf/icml/SonKKHY19}. This work aims to learn a general value factorization without any structural constraints by transforming the optimal value function into one which is easily factorizable. 

\noindent $\bullet$ Reference \cite{DBLP:conf/nips/MahajanRSW19}.This work extends QMIX and other value factorization methods by using a hierarchical policy to guide committed and temporally extended exploration.

\noindent $\bullet$ Reference \cite{DBLP:conf/icra/MaLM21}. This work formalizes the multi-agent system as a graph and lets agents communicate with neighbors via graph convolution to solve the multi-agent pathfinding task.

\noindent $\bullet$ Reference \cite{DBLP:conf/aaai/LiuWHHC020}. This work uses a two-stage attention network to estimate whether two agents should communicate and the importance of that communication instance.

\noindent $\bullet\bullet$  Reference \cite{DBLP:journals/corr/abs-2202-03634}. This work incorporates relies on a conventional coupled planner to guide the learning of the communication topology in multi-agent pathfinding.

\noindent $\bullet\bullet$  Reference \cite{DBLP:journals/ijrr/FanLLP20}. This work presents a deep RL-based decentralized collision-avoidance framework for multi-robot path planning based on sensor inputs, with numerical and experimental validation results.

\noindent $\bullet$  Reference \cite{DBLP:journals/ral/SartorettiKSWKK19}. This work introduces a scalable framework for multi-agent pathfinding which utilizes RL and imitation learning to learn decentralized policies that can scale to more than a thousand agents.

\bmhead{Supplementary information}

Not applicable.

\bmhead{Acknowledgments}

Not applicable.

\section*{Declarations}

\bmhead{Funding}
This work was supported by the Singapore Ministry of Education Academic Research Fund Tier 1.

\bmhead{Conflict of interest/Competing interests}
The authors declare that they have no conflict of interest nor competing interests.

\bmhead{Human and Animal Rights}
This article does not contain any studies with human or animal subjects performed by any
of the authors.

\bmhead{Ethics approval} Not applicable.
\bmhead{Consent to participate} Not applicable.
\bmhead{Consent for publication} Not applicable.
\bmhead{Availability of data and materials} Not applicable.
\bmhead{Code availability} Not applicable.

\bmhead{Authors' contributions}
All authors contributed to the study conception, study design, original draft preparation, as well as review and editing.
Yutong Wang performed figure design, and the literature search and data analysis for communication learning methods, challenges, and benchmarks, as well as for open avenues for research.
Mehul Damani performed the literature search and data analysis for background, as well as communication-free cooperation methods and challenges.
Pamela Wang co-performed the literature search and data analysis for communication learning and benchmarks and Figure~\ref{fig:AI_benchmarks}.
Yuhong Cao performed the literature search and data analysis for multi-robot applications and Figure~\ref{fig:multi-robot path finding}.
Guillaume Sartoretti performed project administration, supervision, and was involved in the literature search for all aspects of this survey.

\begin{appendices}
\label{secA1}

\onecolumn

\begin{longtable}{ |m{2.4cm}|m{1.3cm}|m{0.9cm}|m{7.3cm}|m{2.05cm}|   }
\hline
\multicolumn{5}{|c|}{Representative Works in Cooperative MARL} \\
\hline
\hfil Paper & Learning Setup & Coop. Technique & \hfil Summary & Challenges Addressed \\
\hline
Lauer at al.~\cite{DBLP:conf/icml/LauerR00}&IND&IL& Update Q-Values only if there is guaranteed improvement & NST  \\[0.4cm]
Hysteretic-DQN~\cite{DBLP:conf/iros/MatignonLF07} & IND  & IL & Use smaller learning rate for negative Q-updates & NST \\[0.3cm]
Lenient-DQN~\cite{DBLP:conf/atal/PanaitSL06}&IND&IL& Show progressively decreasing lenience in negative Q-updates & NST\\[0.4cm] 
Stabilising ER~\cite{DBLP:conf/icml/FoersterNFATKW17} & IND & IL & Use importance sampling and fingerprints to stabilize experience replay in MARL & NST   \\[0.4cm] 
MADDPG~\cite{DBLP:conf/nips/LoweWTHAM17}& CTDE & CC & Train a centralized critic for each agent with augmented information & NST \\[0.4cm] 
COMA~\cite{DBLP:conf/aaai/FoersterFANW18} & CTDE & CC & Use counterfactual baseline computed using centralized critic for credit assignment& CA,NST  \\[0.3cm] 
Actor-Attention Critic~\cite{DBLP:conf/icml/IqbalS19} & CTDE & CC & Use an attention mechanism on the centralized critic to dynamically select relevant information & SCA, NST \\[0.4cm] 
VDN~\cite{DBLP:conf/atal/SunehagLGCZJLSL18} & CTDE  & VF & Decompose team value function into a sum of agent-wise value functions & CA,NST \\[0.4cm] 
QMIX~\cite{DBLP:conf/icml/RashidSWFFW18} & CTDE  & VF & Decompose team value function into a non-linear combination of agent-wise value functions using a mixing network & CA,NST \\[0.6cm] 
Omidshafiei et al.~\cite{DBLP:conf/icml/OmidshafieiPAHV17} & CTDE  & IL & Use hysteretic learning, concurrent experience replay, recurrent networks and distillation to learn a multi-task MARL policy  & PO,NST,SCA\\[0.6cm] 
Gupta et al.~\cite{DBLP:conf/atal/GuptaEK17} & CTDE & IL & Use parameter sharing among homogeneous agents to learn more efficiently & SCA \\[0.4cm] 
CommNet \cite{DBLP:conf/nips/SukhbaatarSF16} & IND & CL & Sum continuous messages between different layers of all agents' networks & NST,PO \\[0.4cm] 
BicNet ~\cite{DBLP:journals/corr/PengYWYTLW17} & IND & CL &  Use bi-directional recurrent neural network in the actor-critic paradigm  & NST,PO \\[0.4cm] 
IC3Net ~\cite{DBLP:conf/iclr/SinghJS19} & IND & CL & Selectively communicate through a gating mechanism & NST,PO,SCA \\[0.4cm] 
SchedNet ~\cite{DBLP:conf/iclr/KimMHKLSY19} & CTDE & CL & Selectively broadcast continuous messages by learning importance of each agent’s observation & NST,PO \\
\hline
\caption{
Summary of the major cooperative MARL algorithms reviewed in this paper.
Learning setup (Sect.~\ref{learning-setup}) is either independent \textbf{(IND)} or centralized training decentralized execution \textbf{(CTDE)}.
Cooperation techniques are classified into four based on Sect.~\ref{sec:cooperation}, namely independent learners \textbf{(IL)}, centralized critic \textbf{(CC)}, value factorization \textbf{(VF)} and communication \textbf{(CL)}. 
The major challenges in MARL (Sect.~\ref{challeges}) addressed by these algorithms are non-stationarity \textbf{(NST)}, credit assignment \textbf{(CA)}, scalability \textbf{(SCA)}, and partial observability \textbf{(PO)}. 
}
\label{table:allcoop}
\end{longtable}

{\small
\begin{longtable}{|c|c|c|c|c|c|}
\hline
Reference & \begin{tabular}[c]{@{}c@{}}Scope of \\ communication\end{tabular} & Time-lagged & Setting & Shared Information \\[0.3cm]
\hline
AC-CNet~\cite{DBLP:journals/corr/MaoGNLWKMSX17} & Global & No & Co & Observation \\
A-CCNet~\cite{DBLP:journals/corr/MaoGNLWKMSX17} & Global & No & Co & Observation \& Action \\
IS~\cite{DBLP:conf/iclr/KimPS21} & Global & Yes & Co & Intention \\
CCOMA~\cite{DBLP:journals/corr/abs-2004-00470} & Partial & No & Co & Observation \\
VBC~\cite{DBLP:conf/nips/ZhangZL19} & Targeted & No & Co & Observation \\
ATOC~\cite{DBLP:conf/nips/JiangL18} & Targeted & No & Co & Observation \& Intention \\
DGN~\cite{DBLP:conf/iclr/JiangDHL20} & Partial & No & Co & Observation \\
DHC~\cite{DBLP:conf/icra/MaLM21} & Partial & No & Co & Observation \\
DCC~\cite{DBLP:journals/corr/abs-2109-05413} & Targeted \& Partial & No & Co & 
\begin{tabular}[c]{@{}c@{}}Observation \& \\ Relative position\end{tabular} \\
CommNet~\cite{DBLP:conf/nips/SukhbaatarSF16} & Global & No & Co & Observation \\
SchedNet~\cite{DBLP:conf/iclr/KimMHKLSY19} & Targeted & No & Co & Observation \\
Agarwal et.al.~\cite{DBLP:conf/atal/AgarwalKSL20} & \begin{tabular}[c]{@{}c@{}}Global (unrestricted) \\ \& Partial (restricted)\end{tabular} & No & Co & \begin{tabular}[c]{@{}c@{}}Observation \& \\ Environment Information\end{tabular} \\
IC3Net~\cite{DBLP:conf/iclr/SinghJS19} & Targeted & Yes & Co \& Mix \& Com & Observation \\
GA-Comm~\cite{DBLP:conf/aaai/LiuWHHC020} & Targeted & No & Co & Observation \\
GA-AC~\cite{DBLP:conf/aaai/LiuWHHC020} & Targeted & No & Co & Observation \& Action \\
BicNet~\cite{DBLP:journals/corr/PengYWYTLW17} & Global & No & Co & Observation \& Action \\
MS-MARL~\cite{DBLP:journals/corr/abs-1712-07305} & Global & No & Co & Intention \\
TarMAC~\cite{DBLP:conf/icml/DasGRBPRP19} & Targeted & Yes & Co \& Mix \& Com & Observation \\
\begin{tabular}[c]{@{}c@{}} Blumenkamp \\ and Prorok\end{tabular}
\cite{DBLP:conf/corl/BlumenkampP20} & Partial & No & Co \& Mix & Observation \\
MD-MADDPG~\cite{DBLP:journals/ml/PesceM20} & Global & No & Co & Observation \\
FlowComm~\cite{DBLP:conf/atal/DuLMLRWCZ21} & Targeted & No & Co & Observation \\
DIAL~\cite{DBLP:conf/nips/FoersterAFW16} & Global & Yes & Co & Observation \\ 
RIAL~\cite{DBLP:conf/nips/FoersterAFW16} & Global & Yes & Co & Observation \\ 
PICO~\cite{DBLP:journals/corr/abs-2202-03634} & Partial \& Targeted & No & Co & Observation \\
FCMNet~\cite{DBLP:journals/corr/abs-2201-11994} & Global & No & Co & Observation \\
Cao et al. ~\cite{DBLP:conf/iclr/CaoLLLTC18}  & Global  & No  & Mix  & Intention \& Proposal   \\
Noukhovitch et al. ~\cite{DBLP:conf/atal/NoukhovitchLLC21}   & Global  & No  & Mix  & Proposal  \\
Mihai et al. ~\cite{DBLP:journals/corr/abs-2106-02067}   & Global  & No  & Co  & Observation  \\
Li et al. ~\cite{DBLP:conf/nips/LiB19}  & Global  & No  & Co  & Observation  \\
Freed et al. ~\cite{DBLP:conf/iros/FreedJSC20}   & Global  & No  & Co  & Observation  \\
ForMIC (implicit)~\cite{DBLP:journals/ral/ShawWWS22} & Partial & Yes & Co & Memory \\
\hline 
\caption{
Summary of the communication learning algorithms reviewed in this paper and detailed in Section~\ref{sec:comms-learning}.
Communications can be classified as either explicit or implicit.
Implicit communications are defined as indirect communication between agents in which agents transmit information via their shared environment.
In contrast, explicit communications are separated from the environment, and let agents directly send messages to other agents (All algorithms in the table except ForMIC consider explicit communication).
The scope of communication can either be global, partial, or targeted.
Global communication refers to settings where agents broadcast messages to all other agents.
In partial communications, agents only communicate to other agents within a certain range.
Finally, targeted communication means that agents (learn to) select the specific agents to communicate/listen to at each time-step.
Time-lagged communications consider cases where agents use messages received at one time-step, as input to their decision-making at the next time-step.
Tasks setting include: cooperative  \textbf{(Co)}, mixed  \textbf{(Mix)}, and competitive  \textbf{(Com)}.
Shared information refers to the high-level information encoded within messages shared.
}
\label{table:comm}
\end{longtable}}

\end{appendices}

\begin{multicols}{2}

\bibliographystyle{sn-vancouver}
\bibliography{CRRR-Survey2022}

\end{multicols}
\end{document}